\documentclass{article}

\usepackage{PRIMEarxiv}

\usepackage[utf8]{inputenc} 
\usepackage[T1]{fontenc}    
\usepackage{hyperref}       
\usepackage{url}            
\usepackage{booktabs}       
\usepackage{amsfonts}       
\usepackage{nicefrac}       
\usepackage{microtype}      
\usepackage{lipsum}
\usepackage{fancyhdr}       
\usepackage{graphicx}       
\graphicspath{{media/}}     

\usepackage{amssymb}
\usepackage{amsmath}

\usepackage{mathtools}
\usepackage[table]{xcolor}
\usepackage[belowskip=-5pt,aboveskip=2pt]{caption}

\title{ASTra: A Novel Algorithm-Level Approach to Imbalanced Classification
}

\author{
  David Twomey, Denise Gorse \\
  Department of Computer Science \\
  University College London \\
  London\\
  \texttt{\{d.twomey, d.gorse\}@cs.ucl.ac.uk} \\
}

\begin{document}
\maketitle

\begin{abstract}
We propose a novel output layer activation function, which we name ASTra (Asymmetric Sigmoid Transfer function), which makes the classification of minority examples, in scenarios of high imbalance, more tractable. We combine this with a loss function that helps to effectively target minority misclassification. These two methods can be used together or separately, with their combination recommended for the most severely imbalanced cases. The proposed approach is tested on datasets with IRs from 588.24 to 4000 and very few minority examples (in some datasets, as few as five). Results using neural networks with from two to 12 hidden units are demonstrated to be comparable to, or better than, equivalent results obtained in a recent study that deployed a wide range of complex, hybrid data-level ensemble classifiers.
\end{abstract}

\keywords{Classification \and Class imbalance \and Adaptive Activation Function \and Asymmetric sigmoid \and Confusion matrix \and Geometric mean}

\section{Introduction}
This paper addresses the challenge of handling extreme class imbalance, defined here as a situation in which negative examples, conventionally the majority, outnumber positive examples, usually the ones of most interest, by a factor of 500 or more (in other words, have an imbalance ratio (IR)$\geq$ 500). Such problems are not in fact uncommon, and arise in application areas such as fraud detection~\cite{ref1} and cheminformatics~\cite{ref2}. We make use of two methods, that tackle different, but complementary, aspects of the class imbalance problem:
\begin{itemize}
\item ASTra, a novel, adaptive, asymmetric output layer activation function, which makes the correct classification of minority examples easier.
\item A loss function based on an approximated confusion matrix, which aggressively targets the misclassification of minority examples.
\end{itemize}
Our proposed methods have the advantage of being easy to implement and integrate into the workflow of any model that makes binary predictions normally generated by a sigmoid activation (transfer) function. In addition, the paper presents a new means of monitoring training and validation performance, especially valuable in cases of high class imbalance, that could potentially be used with any training regime, independently of the proposed methods.

\section{Background and Related Work}
\label{sec:background}

\subsection{The Confusion Matrix and Its Associated Metrics}

For a binary problem, in which columns refer to the prediction labels and rows to the actual labels (targets), the confusion matrix (CM) can be defined by 

\begin{equation}
CM =
  \begin{pmatrix}
    TN&FP\\
    FN&TP
  \end{pmatrix},
\end{equation}
in which $TP$ denotes the \emph{true positives} (correctly predicted positive examples), $TN$ denotes the \emph{true negatives} (correctly predicted negative examples), $FN$ denotes the \emph{false negatives} (positive instances predicted to be negative), and $FP$ the \emph{false positives} (negative instances predicted to be positive).

The most complete descriptor of classification performance that can be derived from the confusion matrix is the matrix itself. However, many summary statistics have been proposed (discussed, for example, in~\cite{ref3}), though some of these are of considerably more value than others for imbalanced data. For such data, where conventionally the minority class is labelled positive, false negatives are the most problematic issue: if all of the small number of positive instances are predicted wrongly, the model is effectively useless. However, the accuracy (proportion of correct predictions) may even so be deceptively very high, making accuracy the worst-choice performance statistic to quote for imbalanced data.

Conversely, the \emph{Matthews Correlation Coefficient} (MCC)~\cite{ref4},
\begin{equation}
MCC = \frac{TP \times TN - FP \times FN}
{\sqrt{(TP + FP)(TP + FN)(TN + FP)(TN + FN)}},
\end{equation}
is considered to be a `gold standard' for measuring performance for imbalanced datasets~\cite{ref5}, and hence we choose to use this measure in our results reporting.

Additionally to the MCC, we make use of the \emph{geometric mean} (G-Mean) of the \emph{true positive rate} (sensitivity) $TPR=TP/(TP+FN)$ and \emph{true negative rate} (specificity) $TNR=TN/(TN+FP)$,
\begin{equation} \label{eq:g-mean}
    G\text{-}Mean = \sqrt{TPR\times TNR},
\end{equation}
both as a performance measure, for comparison with the results of~\cite{ref6}, and to build
a loss function. This \emph{GMN loss function}, whose construction is outlined in section 3.2, was proposed by us in~\cite{ref7}, and is used in the current work due to its effectiveness in targeting false negatives, these being the most intractable element of the classification problem for datasets with a high IR.

\subsection{Related Work}

Methods for addressing class imbalance can be broadly grouped into three categories: data-level techniques, algorithm-level methods, and hybrid approaches~\cite{ref8}. \emph{Data-level} techniques aim to reduce the level of class imbalance through some form of sampling, for example oversampling the minority or undersampling the majority class. However, oversampling can lead to overfitting, and undersampling to a loss of important information~\cite{ref9}. \emph{Algorithm-level} methods, commonly implemented with a weight or cost scheme, involve modifying the underlying learner or its output in order to reduce bias toward the majority class. In cost-sensitive learning, penalties are assigned to each class through a cost matrix, though its definition often needs to be done empirically, or using expert domain knowledge~\cite{ref8}. \emph{Hybrid} approaches combine data-level and algorithm-level approaches in various ways to handle the class imbalance problem. Examples include RUSBoost~\cite{ref10} and the recently-proposed HD-Ensemble method of~\cite{ref6}.

An alternative algorithm-level approach, taken here, is to replace a loss function based on distance of an output from its target with one based on classification success. For imbalanced datasets, such a loss function may be able to avoid the trap of decreasing the loss by merely driving already-correct majority outputs closer and closer to their target values. The earliest work in this area, to our knowledge, dates back to 2013, when~\cite{ref11} proposed a loss function based on the F1 score and used this for the enhancement of document images. Two works appeared in 2017, the current authors' proposal of an approximated G-Mean~\cite{ref7}, and a new F1 score based loss function~\cite{ref12}. In 2019, two further variants of the F1 score were proposed, the first~\cite{ref13} used to train a CNN to classify emotions in tweets, and the second~\cite{ref14} a proposal specifically for linear models, applied to synthetic and image data. Also in 2019,~\cite{ref15} used a CNN and a multi-class variant of the F1 score to perform cell segmentation. Most recently, in 2021,~\cite{ref16} used a CNN, trained this time with an approximated MCC, to classify skin lesions.

Notably, most of the above work used loss functions derived from the F1 score. However, this measure has been criticised for insufficiently addressing false negatives~\cite{ref17}, and may thus not be optimal in areas such as medical diagnosis. The MCC may also, as demonstrated in section \ref{astra_function}, excessively emphasise false positives at the expense of false negatives. The G-Mean of Eq. (\ref{eq:g-mean}), however, aggressively targets false negatives due to its product form, which is the reason we choose to use our approximated G-Mean loss~\cite{ref7} in this current work.

\section{Proposed Methods}

\subsection{The ASTra (Asymmetric Sigmoid Transfer) Function} \label{astra_function}

ASTra is an asymmetric sigmoid-type activation function, recommended for use in the final network layer only, that is designed to facilitate the classification of minority examples by allocating a larger proportion of the output range to these examples. Its form is derived from that of the Richards growth equation~\cite{ref18}, used to model vegetation growth and the growth of young mammals and birds, and (assuming a conventional minority-1) is defined by

\begin{equation}
    ASTra(x,b)=1-(1+be^{bx})^{-\frac{1}{b}}.
\end{equation}
The gradient of the ASTra transform is maximum at $x = 0$, which point we define as the threshold, $\tau $, given by
\begin{equation}
    \tau (b)=1-(1+b)^{-\frac{1}{b}}.
\end{equation}
When $b = 1$, the form of the standard sigmoid can be recovered, while for $b > 1$ lower thresholds $\tau  < 0.5$  make the classification of target-1 examples, the assumed minority, easier. $y(x,b)$ and its first derivative are shown in Figs. 1(a), (b), for $\tau = 0.5,0.25,0.05$, noting that we have so far found 0.05 to be the lowest workable value for $\tau$ (attempts to compress all majority examples into $< 5\%$ of the output range having led to numerical instabilities).

\begin{figure}
\includegraphics[width=\textwidth]{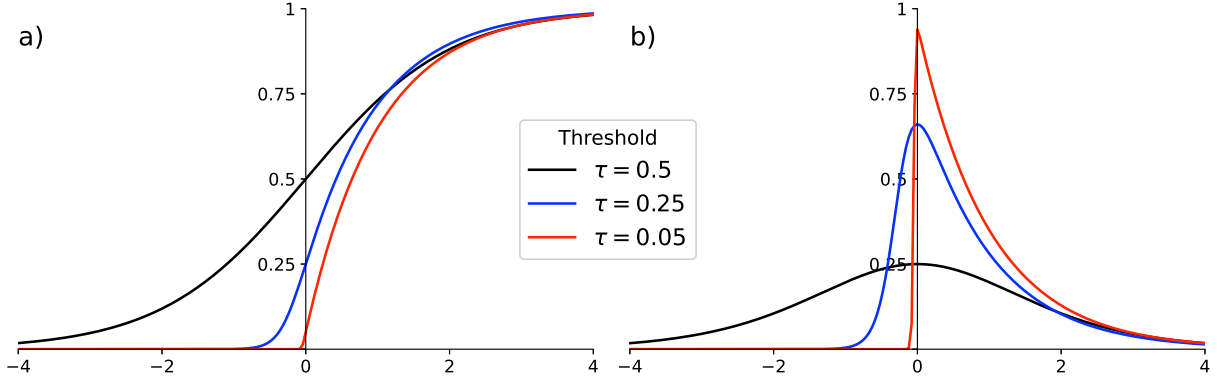}
\caption{(a) Activation function $ASTra(x,b)$ and (b) its first derivative, for $\tau (b)=0.5, 0.25, 0.05$.} 
\label{fig1}
\end{figure}

\textbf{Adaptation of Loss Functions to Accommodate ASTra.} At first sight, this looks easy: simply replace the usual sigmoid output
by Eq. (4). However, it is not always so straightforward, as an examination of ASTra's effect of the computation of a binary cross-entropy (BCE) loss will show.

Consider the contribution 
$J_{BCE}(\hat{y},y) = -y\log\hat{y} - (1-y)\log(1-\hat{y})$ 
of a single example, with prediction $\hat{y} = ASTra(x, b)$ and target $y\in\{0,1\}$, to the train set loss. As is apparent in Fig. 2(a), for $b = 1 $ ($\tau = 0.5$), for any input $x> 0.0$, $J_{BCE}(\hat{y},0) > J_{BCE}(\hat{y},1)$, correctly, since $\hat{y}$ will be on the `correct side' of the threshold and so closer to target 1 than to 0, with the converse also correctly being true for inputs $x < 0.0$. However, for $b>1$ this is not true for all values of the input $x$; there will be a range of $x$, specifically, $0 < x < (\log[(2^{b}-1) / b]) / b$, within which the loss contributions are wrongly ordered. Fig. 2(b) shows an example of this for $\tau=0.25$ ($b=7.396$); within the shaded band it is the case that $J_{BCE}(\hat{y},0) < J_{BCE}(\hat{y},1)$, so outputs which are `more right than wrong' are instead being informed that the opposite is the case.

The fundamental problem is that currently $J_{BCE}(\hat{y},0)$ and $J_{BCE}(\hat{y},0)$ cross at 0.5, not at $\tau$, and the solution is to use within the BCE loss not $\hat{y}$ but a transform $z(\hat{y},\tau)$, given by

\begin{equation}
    z(\hat{y},\tau) = \frac{\hat{y}(1-\tau)}{\hat{y}(1-\tau)+(1-\hat{y})\tau},
\end{equation}
such that $z$, $1-z$ will now cross at $\tau$, as can be seen in Fig. 3, for $\tau=0.5, 0.25$, and 0.05, and hence the loss contributions $J_{BCE}(z,0)$ and $J_{BCE}(z,1)$ will do so, also.
For some loss functions, such as mean squared error, it is unnecessary to transform $\hat{y}$ in this way. However, the $z$-transform should be carried out for any classification loss function that implicitly pivots around a threshold of 0.5, including all loss functions, such as GMN, derived from the confusion matrix. 

\begin{figure}
\includegraphics[width=\textwidth]{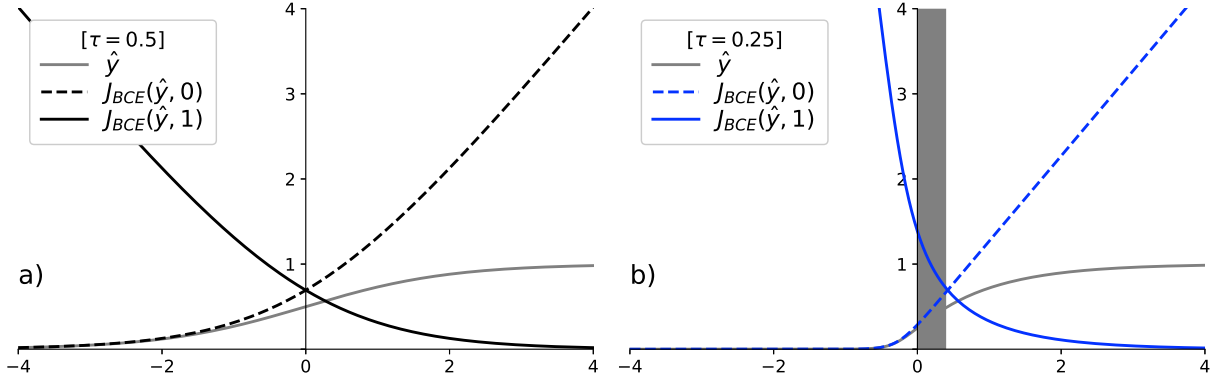}
\caption{Loss functions for targets 0 and 1 for (a) $\tau(b)=0.5$ and (b) $\tau(b)=0.25$.} 
\label{fig2}
\end{figure}

\begin{figure}
\includegraphics[width=6.5cm]{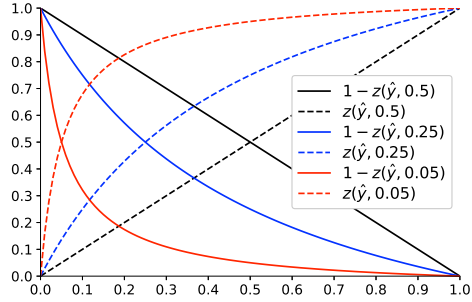} \centering
\caption{$z$-transforms for $\tau=0.5, 0.25, 0.05$.} 
\label{fig3}
\end{figure}

\textbf{Learning the Threshold Value.} It would in principle be possible to use a fixed value of $\tau$ and set this via hyperparameter optimisation. However, in experiments so far, it has appeared that a fixed $\tau < 0.5$ during weights-learning is non-optimal, and that it is better to begin with a somewhat larger threshold value that then decreases during the network's learning process. There is necessarily therefore a question of how $\tau$ should be decreased. While it would be possible to experiment with a schedule of decrease, for example, exponential or linear, a more attractive option is to learn the value of the underlying `slope' parameter, $b$, alongside the network's weights.

For a conventional minority target of 1, the relevant values of $b$ should be constrained to be $\geq 1$(corresponding to a threshold $\tau \leq 0.5$). Given this, it is useful to express $b$ in terms of an unconstrained underlying parameter $\beta$: 

\begin{equation}
b = \begin{cases}
2 + \beta & \beta > 0\\
1 + e^{\beta } &\text{otherwise}
\end{cases}.
\end{equation}
It is preferable to begin with $b(\beta)$ in its linear range, i.e., $b \geq 2$, so that the threshold can change sufficiently quickly at the start of the learning process. We have used a value of $\tau_{init} = 0.25$ ($b = 7.396$), around halfway between the achievable threshold maximum of 0.5 and the recommended (on stability grounds) minimum of 0.05, but clearly this is a value that could be treated as a hyperparameter and optimised.

\subsection{The GMN Loss Function}

The construction of this loss function, first presented by us in~\cite{ref7}, begins with an approximated confusion matrix,
\begin{equation}
    CM_{apx} = 
    \begin{pmatrix*}
    TN_{apx}&\phantom{-}FP_{apx}\\
    FN_{apx}&\phantom{-}TP_{apx}
  \end{pmatrix*}
   = 
   \begin{pmatrix*}[l]
    \sum_{i}^{n}{(1 -\hat{y_i})(1-{y_i})} &
    \phantom{-}\sum_{i}^{n}{\hat{y_i}(1-y_i)} \\
    \sum_{i}^{n}{(1 - \hat{y_i})y_i} &
    \phantom{-}\sum_{i}^{n}{\hat{y_i}y_i}
  \end{pmatrix*},
\end{equation}
in which $\hat{y_i}$ and $y_i$ are the predicted and target values for the $i^{th}$ example, and the sums are over the set of examples for which the approximated matrix is required. From $CM_{apx}$ one can then generate the necessary approximation to the G-Mean, and our loss function
\begin{equation}
J_{GMN} = 1 - G\text{-}Mean_{apx} = 1- \sqrt{\frac{\sum_{i}{(1 - \hat{y_i})(1-y_i)} \cdot  \sum_{i}{\hat{y_i}y_i}}{m_0 \cdot m_1 }},
\end{equation}
in which $m_0$, $m_1$  are the numbers of negative, positive examples, respectively. The necessary derivatives with respect to network outputs can then be either derived algebraically or obtained via automatic differentiation.\footnote{All derivatives, including those associated with the use of the ASTra transform, may be obtained from the authors on request; space precludes their inclusion here.}

\section{Data, Experimental Process, and Performance Measurement}

\subsection{Data}

There are not many easily available datasets with the high level of imbalance required for this work. However, a recent study~\cite{ref6} has looked at data with IRs ranging from 9.08 to 970.6, with data made available online~\cite{ref19}. It was also useful that these data were used in~\cite{ref6} to compare HD-Ensemble, a high-performing hybrid, data-level ensemble classifier devised by that study's authors, to seven other state-of-the-art ensemble classifiers, including RUSBoost~\cite{ref10}. The three most imbalanced datasets from~\cite{ref6}, named there as \emph{skinnonskin}, \emph{cod-rna}, and \emph{ijcnn1} (IR = 970.6) were downloaded from~\cite{ref19}.\footnote{Modified, where necessary, to record target-1 for the minority class and target-0 for the majority, a requirement for the application of the ASTra activation function.}  In addition to the original files, the two largest datasets from~\cite{ref19} were minority-undersampled to create more challenging  problems (IRs of 3500, 4000, with one positive example in each of five folds), extracting five positive samples randomly at each run, to eliminate bias created by `good' or `bad' subset picks.

\begin{table}
\caption{Datasets used in this work. The shaded bars correspond to the original datasets of~\cite{ref6} and the unshaded bars to our additional minority-undersampled datasets.}\label{tab1}
\centering
\setlength{\tabcolsep}{11pt} 
\renewcommand{\arraystretch}{1.05} 
\begin{tabular}{lrlrrr}
\hline\hline
\textbf{Dataset} & \textbf{IR} & \textbf{Abb.} & \boldmath${n_x}$ & \boldmath${m_{tot}}$ & \boldmath${m_1}$ \\
\hline
\cellcolor{gray!15}\emph{skinnonskin} & \cellcolor{gray!15}588.24 & \cellcolor{gray!15}\emph{skin-588} & \cellcolor{gray!15}3 & \cellcolor{gray!15}20034 & \cellcolor{gray!15}34 \\
above, undersampled & 4000 & \emph{skin-4000} & 3 & 20005 & 5 \\
\cellcolor{gray!15}\emph{cod-rna} & \cellcolor{gray!15}763.27 & \cellcolor{gray!15}\emph{cod-763} & \cellcolor{gray!15}8 & \cellcolor{gray!15}19871 & \cellcolor{gray!15}26 \\
above, undersampled & 3500 & \emph{cod-3500} & 8 & 17505 & 5 \\
\cellcolor{gray!15}\emph{ijcnn1} & \cellcolor{gray!15}970.60 & \cellcolor{gray!15}\emph{ijcnn-971} & \cellcolor{gray!15}22 & \cellcolor{gray!15}4858 & \cellcolor{gray!15}5 \\
\hline
\end{tabular}
\end{table}

The properties of the datasets used here are summarised in Table 1, in which $n_x$ is the number of input features, $m_{tot}$ the total number of examples in the dataset, and $m_1$ the number of minority-positive examples. It should be noted that while we followed~\cite{ref6} in creating stratified test folds, we used $10\times5$CV testing rather than $5\times10$CV, using three folds for training, and one each for validation and testing, as it was unclear how tenfold stratified sampling had been done in~\cite{ref6} in the case of \emph{ijcnn1}, with only five positive examples.

\subsection{Details of the Experimental Process}
The focus of this work is not the test problems of Table 1, per se, but a comparison of methodologies, and therefore simple choices were made for network architectures, with no attempt to optimise learning rates, etc., in order to get the best solution in each case. For each test problem, four candidates were compared: training with a binary cross-entropy loss (referred to as BCE); training with a binary cross-entropy loss and ASTra transform (BCE-ASTra); training with the GMN loss function (GMN); and training with the GMN loss function and ASTra transform (GMN-ASTra). The architecture for each test problem was a neural network with one hidden layer of Leaky ReLU neurons (negative slope coefficient of 0.3), with $n_h = ceil((n_x + n_y)/2)$, and a single ASTra output (the standard sigmoid being recoverable from this when $b = 1$). Network weights used He initialisation in the hidden layer and Glorot initialisation in the output layer. All models were trained using Adam, with a training rate of $\eta = 0.001$  for the network weights.
$\eta_b$, the training rate associated with the slope parameter $b$, was set according to the adaptive rule of Eq. (10),
\begin{equation}
\eta_b = \begin{cases}
\min({k_{mult} \cdot \eta_b, \eta_{bmax}}) & \text{if \emph{e-ratio} $>$ 1.0}\\
\max({k_{dec} \cdot \eta_b, \eta_{bmin}}) & \text{if \emph{e-ratio} $<$ 1.0}\\
\eta_b &\text{otherwise}
\end{cases},
\end{equation}
the objective of which was to allow the threshold adjustment to proceed more quickly when the model is struggling to handle its train set class imbalance, but slow down when conditions are less taxing, as measured by the \emph{e-ratio},
\begin{equation}
e\text{-}ratio = FNR_{apx} / FPR_{apx},
\end{equation}
which measures the difficulty of classifying positive (minority) examples with respect to negative ones.
We here chose  $\eta_{bmin}$  (also the starting value of $\eta_b$) to be $0.01$, $\eta_{bmax}=0.5$, $k_{mult}=1.1$, and $k_{dec}=0.99$. These values were not optimised, in line with our focus on comparison of training methods, rather than on outright performance.

The one form of problem-specific optimisation we performed was the extraction of the best weights (and $b$ value, where relevant) with respect to a validation set.  However, it was unfeasible to use early stopping as best validation performance could in some cases occur late, after an early setback. Therefore, we trained for a fixed  10,000 epochs, extracting a `best-on-val' set of parameters, for whatever epoch this occurred, with respect to the chosen validation performance measure ($FNR_{apx}$, for the reasons discussed in the following section).

\subsection{Performance Measurement}

Two aspects of performance measurement need to be considered, that which is carried out during the training process, and that which is used to assess the results on the test sets. For the latter, we use the G-Mean, as it is the basis of one of our considered loss functions, and one of the two performance measures quoted in~\cite{ref6}. However, we do not use AUROC, as in~\cite{ref6}, as this metric has been criticised in the case of imbalanced data~\cite{ref20}, and we found it to distinguish very poorly between the methods of this paper. We substitute the MCC as our second measure, due to its high regard as a performance metric for imbalanced data.

Turning to in-training performance monitoring, we introduce an innovation in that we retain the explanatory value, for this purpose, of confusion matrix based metrics, but exchange the usual, counting-measure forms for ones based on the approximated matrix—i.e., we use standard formulae for G-Mean, etc., but base the calculation on the matrix elements of (4) rather than (1). Metrics derived from $CM_{apx}$ re-introduce the `by how much?' that counting-measure classification metrics lose. This is particularly important when monitoring validation performance, for which, in this work, we select to monitor $FNR_{apx}$, the approximated false negative rate, rather than $MCC_{apx}$, for example, as in these extreme scenarios it is test set \emph{FNR} that largely determines classification (in terms of MCC, G-Mean) success. The other element of in-training performance monitoring relates to the train set itself, rather than to validation. We record the \emph{e-ratio} of Eq. (11)
for the train set as a way to track the difficulty a model is having with the train set class imbalance; this ratio is of interest in understanding how the methods deal with class imbalance, with a large difference being discovered between BCE loss and GMN loss training, as will be seen later.

\section{Results}

The results of the $10\times5$CV study are summarised in Table 2, below, in which averages and standard deviations (in brackets) are quoted for both G-Mean and MCC, and where BCE, GMN refer to the loss functions ($J_{BCE}$, $J_{GMN}$) used.

\begin{table}
\caption{Mean fold-aggregated test performance ($10\times5$CV), with standard deviations bracketed. Cells with bold type and shading denote winners or ties, where to be deemed a winner a method needs to outperform its competitors with $p \leq 0.05$.}\label{tab2}
\centering
\setlength{\tabcolsep}{8pt} 
\renewcommand{\arraystretch}{1.1} 
\begin{tabular}{l c c c c}
\hline\hline
 & \multicolumn{2}{c}{{\bfseries Without ASTra}} & \multicolumn{2}{c}{{\bfseries With ASTra}} \\
& \textbf{BCE} & \textbf{GMN} & \textbf{BCE} & \textbf{GMN} \\
\hline
\multicolumn{5}{l}{{\bfseries \emph{skin-588}}} \\
{ G-Mean} & \cellcolor{gray!15}\textbf{0.892 (0.273)} & \cellcolor{gray!15}\textbf{0.966 (0.051)} & \cellcolor{gray!15}\textbf{0.981 (0.041)} & \cellcolor{gray!15}\textbf{0.946 (0.126)} \\
{ MCC} & \cellcolor{gray!15}\textbf{0.892 (0.273)} & \cellcolor{gray!15}\textbf{0.966 (0.051)} & \cellcolor{gray!15}\textbf{0.976 (0.042)} & 0.763 (0.273) \\
\hline
\multicolumn{5}{l}{{\bfseries \emph{skin-4000}}} \\
{ G-Mean} & \cellcolor{gray!15}\textbf{0.760 (0.436)} & \cellcolor{gray!15}\textbf{0.879 (0.331)} & \cellcolor{gray!15}\textbf{0.800 (0.408)} & \cellcolor{gray!15}\textbf{0.840 (0.374)} \\
{ MCC} & 0.760 (0.436) & \cellcolor{gray!15}\bf{0.843 (0.368)} & \cellcolor{gray!15}\textbf{0.800 (0.408)} & 0.688 (0.382) \\
\hline
\multicolumn{5}{l}{{\bfseries \emph{cod-763}}} \\
{ G-Mean} & 0.449 (0.396) & 0.768 (0.206) & \cellcolor{gray!15}\textbf{0.862 (0.208)} & \cellcolor{gray!15}\textbf{0.843 (0.206)} \\
{ MCC} & 0.433 (0.391) & 0.419 (0.301) & \cellcolor{gray!15}\textbf{0.760 (0.229)} & 0.542 (0.291) \\
\hline
\multicolumn{5}{l}{{\bfseries \emph{cod-3500}}} \\
{ G-Mean} & 0.240 (0.436) & 0.357 (0.486) & \cellcolor{gray!15}\textbf{0.640 (0.490)} & \cellcolor{gray!15}\textbf{0.519 (0.509)} \\
{ MCC} & 0.240 (0.436) & 0.255 (0.397) & \cellcolor{gray!15}\textbf{0.421 (0.369)} & 0.206 (0.225) \\
\hline
\multicolumn{5}{l}{{\bfseries \emph{ijcnn-971}}} \\
{ G-Mean} & 0.000 (0.000) & 0.120 (0.331) & \cellcolor{gray!15}\textbf{0.478 (0.508)} & \cellcolor{gray!15}\textbf{0.596 (0.497)} \\
{ MCC} & -0.000 (0.001) & 0.072 (0.208) & \cellcolor{gray!15}\textbf{0.162 (0.179)} & \cellcolor{gray!15}\textbf{0.179 (0.162)} \\
\hline
\end{tabular}
\end{table}

The large standard deviations, apparent also in the results of~\cite{ref6}, are a consequence of fold-based aggregation of results in a situation where a small number of false negatives (in the case of the \emph{skin-4000}, \emph{cod-3500}, and \emph{ijcnn-971} datasets, a single false negative) on an individual fold can cause its G-Mean or MCC to drop from near-unity to zero. This in turn causes some results, with a sample size of 50 ($10\time 5$CV), to be not statistically significantly distinguishable; e.g., in the case of the \emph{skin-588} dataset, the G-Means of the four methods cannot be separated at $p\leq 0.05$  in spite of an apparent-win by BCE with ASTra (BCE-ASTra). Nevertheless, despite the necessity of a registering a proportion of the results as `ties', a pattern emerges, with the problems forming a hierarchy of difficulty, at each level of which different methods will be appropriate:
\begin{itemize}
\item \emph{skin-588} is an example of a problem for which no additional measures to address IR need to be taken, since vanilla BCE training can do as well (subject to the above note on statistical significance testing) as any of the more advanced methods.
\item \emph{skin-4000}, the minority-undersampled version of \emph{skin-588}, is a more difficult problem, for which BCE training falls behind with respect to MCC. For this problem, adopting the principle that a minimal intervention is to be preferred, GMN training, without the ASTra transform, would be recommended.
\item \emph{cod-763}, \emph{cod-3500}, and \emph{ijcnn-971} are examples of problems that appear to clearly benefit from the ASTra transform. For the \emph{cod} variants it is BCE-ASTra that does best with respect to both G-Mean and MCC, having fewer false positives for these problems, while for \emph{ijcnn-971} BCE-ASTra and GMN-ASTra tie. 
\end{itemize}
The above list does not include a problem for which GMN-ASTra does best. However, we would expect such problems to exist. GMN-ASTra attacks false negatives at the expense of a somewhat increased number of false positives, and if it is possible to avoid this by using BCE-ASTra, that is desirable. But it is likely there are problems that require an increase in false positives to be tolerated in order to have any chance of recognising instances of the minority class (i.e., of having G-Means and MCCs above zero), and in such cases we would expect GMN-ASTra to be preferred.

For the three datasets we have in common with~\cite{ref6} it is possible to draw performance comparisons with that work, though retaining our concerns about the use of AUROC for highly imbalanced data, which we found did not distinguish well between the methods. The two best-performing methods of~\cite{ref6}, in relation to G-Mean and AUROC, were the HD-Ensemble method of~\cite{ref6} and RUSBoost~\cite{ref10}. For \emph{skin-588}, GMN, BCE-ASTra, and GMN-ASTra all had statistically significantly indistinguishable performance from RUSBoost, with respect to both metrics. For \emph{cod-763}, the two ASTra variants tied with RUSBoost on AUROC, but beat RUSBoost on G-Mean, while on \emph{ijcnn-971} they achieved a better G-Mean than state-of-the-art HD-Ensemble. Given especially that we used small, single-layer neural networks, with that used for \emph{skin-588} having only three neurons, we consider these results encouraging.

As a means of illuminating the difference between our four considered methods, Fig. 4 shows the \emph{e-ratio} of section 4.2, for the \emph{skin-4000} (most highly imbalanced) problem. The GMN variants find handling the imbalance much easier, with \emph{e-ratio} values up to five orders of magnitude lower; also, while ASTra is taxing in terms of \emph{e-ratio}, this can be offset by the use of GMN. 

\begin{figure}
\includegraphics[width=9cm]{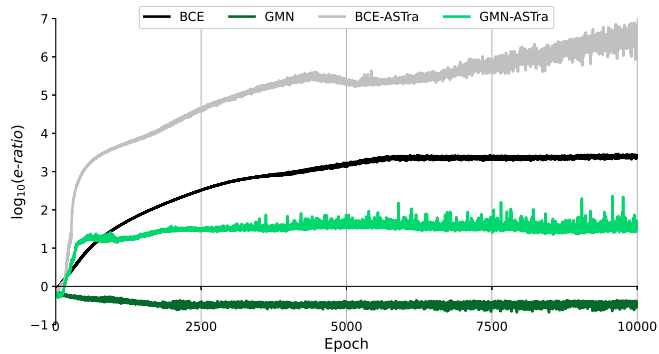}\centering
\caption{Fold-averaged training behaviour for \emph{skin-4000}, in terms of the (logged) \emph{e-ratio} of section 4.2, for each considered method.} \label{fig4}
\end{figure}

\section{Discussion}
This paper has introduced a novel adaptive output layer activation function, which we have named ASTra, that can either in combination with, or separately from, our previously proposed GMN loss~\cite{ref7} work to facilitate the classification of minority examples for extremely high IRs. We additionally introduced a novel use of metrics based on an approximated confusion matrix for performance monitoring during training.
The proposed methods were applied to the most imbalanced datasets in a recent extensive study of ensemble classifiers~\cite{ref6} and achieved  performances comparable to those reported in~\cite{ref6} for RUSBoost~\cite{ref10}, despite our restricting this initial investigation to the use of neural networks with 3–12 neurons in a single hidden layer. We emphasise, however, that for complex problems it is essential to explore the space of model architectures, since even in the case of a high IR the main difficulties may in fact lie in an insufficient model architecture. We also recommend a progressive exploration of the methods, beginning with the GMN loss alone, then advancing to the addition of the ASTra output layer activation function, as needed.
We aim next to apply our methods to challenging real-world imbalanced datasets requiring deep neural network architectures. Additionally, especially given the nature of the derivatives involved in ASTra computations, we are interested in the use of sharpness-aware minimisation~\cite{ref21} to smooth optima and improve generalisation. Finally, we note here that it is our intention to open-source our code and to ensure its integration with popular machine learning libraries (Pytorch and Tensorflow).

\bibliographystyle{unsrt}  
\bibliography{references}

\end{document}